\documentclass[11pt,a4paper]{article}
\usepackage{authblk}
\usepackage[hyperref]{acl2021}
\usepackage{times}
\usepackage{latexsym}
\usepackage{graphicx}
\usepackage{multirow}
\usepackage{threeparttable}

\usepackage{booktabs}
\usepackage[normalem]{ulem}
\useunder{\uline}{\ul}{}

\usepackage{array}
\newcolumntype{P}[1]{>{\centering\arraybackslash}p{#1}}

\usepackage{microtype}

\aclfinalcopy

\title{CoTexT: Multi-task Learning with Code-Text Transformer}

\author[1]{Long Phan}
\author[2]{Hieu Tran}
\author[1]{Daniel Le}
\author[1]{Hieu Nguyen}
\author[1]{James Anibal}
\author[1]{Alec Peltekian}
\author[1]{Yanfang Ye}
\affil[1]{Case Western Reserve University, Ohio, USA}
\affil[2]{University of Science, VNU-HCM, Vietnam}

\affil[ ]{\textit {\{lnp26, yxy1032\}@case.edu}}

\date{}

\begin{document}
\maketitle
\begin{abstract}
We present CoTexT, a pre-trained, transformer-based encoder-decoder  model that learns the representative context between natural language (NL) and programming language (PL). Using self-supervision, CoTexT is pre-trained on large programming language corpora to learn a general understanding of language and code. CoTexT supports downstream NL-PL tasks such as code summarizing/documentation, code generation, defect detection, and code debugging. We train CoTexT on different combinations of available PL corpus including both "bimodal" and "unimodal" data. Here, bimodal data is the combination of text and corresponding code snippets, whereas unimodal data is merely code snippets. We first evaluate CoTexT with multi-task learning: we perform Code Summarization on 6 different programming languages and Code Refinement on both small and medium size featured in the CodeXGLUE dataset. We further conduct extensive experiments to investigate CoTexT on other tasks within the CodeXGlue dataset, including Code Generation and Defect Detection. We consistently achieve SOTA results in these tasks, demonstrating the versatility of our models. 

\end{abstract}

\section{Introduction}
In recent years, pre-trained language models (LM) have played a crucial role in the development of many natural language processing (NLP) systems. Before the emergence of large LMs, traditional word embedding gives each word/token a global representation. Large pre-trained models such as ELMo \cite{DBLP:journals/corr/abs-1802-05365}, GPT \cite{brown2020language}, BERT \cite{DBLP:journals/corr/abs-1810-04805}, and XLNet \cite{yang2020xlnet} can derive contextualized word vector representations from large corpora. These methods can learn generalized representations of language and have significantly improved a broad range of downstream NLP tasks. These LMs make use of learning objectives such as Masked Language Modeling (MLM) \cite{DBLP:journals/corr/abs-1810-04805} where random tokens in a sequence are masked and the model predicts the original tokens to learn the context. The success of  pre-trained models in NLP has created a path for domain-specific pre-trained LMs, such as BioBERT \cite{DBLP:journals/corr/abs-1901-08746} on biomedical text, or TaBERT \cite{yin2020tabert} on NL text and tabular data.

We introduce CoTexT (Code and Text Transfer Transformer), a pre-trained model for both natural language (NL) and programming language (PL) such as Java, Python, Javascript, PHP, etc. CoTexT follows the encoder-decoder architecture proposed by \cite{DBLP:journals/corr/VaswaniSPUJGKP17} with attention mechanisms. We then adapt the model to match T5 framework proposed by \cite{DBLP:journals/corr/abs-1910-10683}. We test CoTexT by performing exhaustive experiments on multi-task learning of multiple programming languages and other related tasks. 

We train CoTexT using large programming language corpora containing multiple programming languages (including Java, Python, JavaScript, Ruby, etc.). Here, we test different combinations of unimodal and bimodal data to produce the best result for each downstream task. We then fine-tune CoTexT on four CodeXGLUE tasks \cite{lu2021codexglue} including CodeSummarization, CodeGeneration, Defect Detection and Code Refinement (small and medium dataset). Results show that we achieve state-of-the-art values for each of the four tasks. We found that CoTexT outperforms current SOTA models such as CodeBERT \cite{feng-etal-2020-codebert} and PLBART \cite{ahmad2021unified}. 

In this paper we offer the following contribution:

\begin{itemize}
    \item Three different versions of CoTexT that achieve state-of-the-art on the CodeXGLUE's CodeSummarization, CodeGeneration, Defect Detection and Code Refinement (small and medium dataset) tasks. We publicize our CoTexT pre-trained checkpoints and related source code available for future studies and improvements.
\end{itemize}

\section{Related Work}
Recent work on domain adaptation of BERT show improvements compared to the general BERT model. BioBERT \cite{Lee_2019} is further trained from BERT\textsubscript{BASE} on biomedical articles such as PubMed abstracts and PMC articles. Similarly, SciBERT \cite{beltagy-etal-2019-scibert} is trained on the full text of biomedical and computer science papers. The experimental results of these models on domain-specific datasets show the enhanced performance compared to BERT\textsubscript{BASE}. 

Relating specfically to our work, CodeBERT is \cite{feng-etal-2020-codebert} trained on bimodal data of NL-PL pairs. This strategy allows CodeBERT to learn general-purpose representations of both natural language and programming language. GraphCodeBERT \cite{guo2021graphcodebert} is an extension of CodeBERT that moves beyond syntactic-level structure and uses data flow in the pre-training stage to capture the semantic-level structure of code. More recently, PLBART \cite{ahmad2020summarization} is a pre-trained sequence-to-sequence model for NL and PL. Through denoising autoencoding, this model can perform well on NL-PL understanding and generation tasks.

\section{CoTexT}
\subsection{Vocabulary} \label{Vocabulary}
Following the example of T5 \cite{DBLP:journals/corr/abs-1910-10683}, we use the Sentence Piece Unsupervised Text Tokenizer proposed by \cite{DBLP:journals/corr/abs-1808-06226}. The Sentence Piece model extracts the sub-words that contain the semantic context of a sequence. We employ Sentence Piece as a vocabulary model for all of our contributed CoTexT models. However, the special tokens used in code (such as "[", "\{", "\$", etc) are out-of-vocab for the SentencePiece model \footnote{https://github.com/google/sentencepiece}. These tokens have a crucial representative context in programming languages. Therefore, to enhance the robustness of the model, we encode all of these missing tokens into a natural language representation during both self-supervised and supervised training.

\begin{figure}
    \centering
    \includegraphics[width=0.5\textwidth,height=4cm,keepaspectratio]{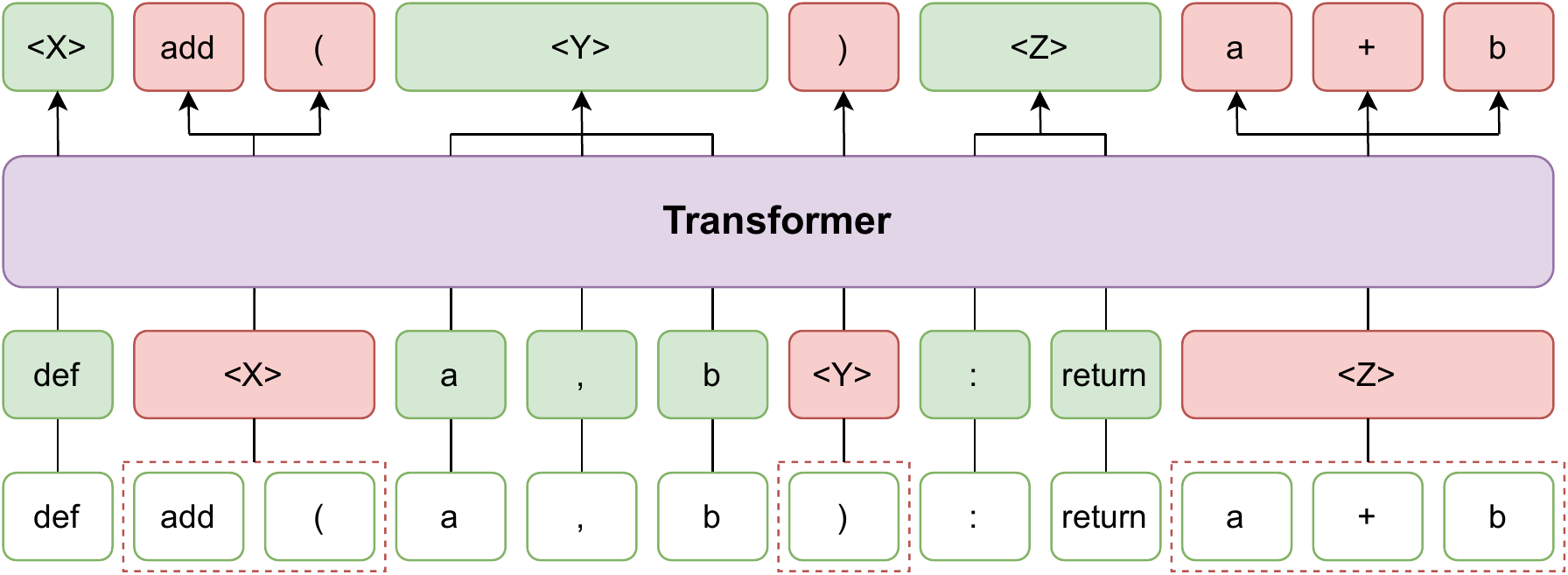}
    \caption{An illustration about Fill-in-the-blank objective}
    \label{fig:model}
\end{figure}

\subsection{Pre-training CoTexT} 
We train CoTexT on both bimodal and unimodal data. Bimodal data contains both code snippets and the corresponding natural text in each sequence, while unimodal data contains only the sequence of code. We use two main datasets during self-supervised training: \textbf{CodeSearchNet Corpus Collection} \cite{husain2020codesearchnet} and \textbf{GitHub Repositories}\footnote{https://console.cloud.google.com/marketplace/details/github/github-repos} data. The combinations of corpus used to train CoTexT are listed in Table \ref{corpus_combination}. To save both time and computing resources, we initialized the checkpoints from the original T5 that was trained on the C4 corpus. \cite{DBLP:journals/corr/abs-1910-10683}.

\subsubsection{CodeSearchNet Corpus Collection}
CodeSearchNet Corpus \cite{husain2020codesearchnet} contains coded functions from open-source non-forked Github repositories. This dataset spans 6 coding languages (Python, Java, Javascript, PHP, Ruby, Go), which facilitates multi-task learning. CodeSearchNet also contains a natural language description for each function. For bimodal data, we simply concatenate the natural language snippet with the corresponding code snippet to create one input sequence. These data are then processed as described in \ref{Vocabulary}.

\begin{table*}[!htbp]
\centering
\caption{Pre-training CoTexT on different combinations of natural language and programming language copora}
\begin{tabular}{@{}lcl@{}}
\toprule
\textbf{Model}               & \textbf{N-modal}                 & \textbf{Corpus combination}                \\ \midrule
T5                       & NL            & C4                                \\
CoTexT (1-CC)     & PL  & \textbf{C}4 + \textbf{C}odeSearchNet                \\
CoTexT (2-CC)   & NL-PL    & \textbf{C}4 + \textbf{C}odeSearchNet                \\
CoTexT (1-CCG) & PL & \textbf{C}4 + \textbf{C}odeSearchNet + \textbf{G}ithub Repos \\ \bottomrule
\end{tabular}
\label{corpus_combination}
\end{table*}

\subsubsection{GitHub repositories}
We download a large collection of Java and Python functions from the GitHub repositories dataset available on Google BigQuery. These Java and Python functions are then extracted and the natural language descriptions are obtained using the pre-processing pipeline from \cite{lachaux2020unsupervised}. These datapoints also run through a pipeline to replace special tokens (as described in \ref{Vocabulary}).
\subsection{Input/Output Representations}

CoTexT converts all NLP problems into a text-to-text format. This means that during both self-supervised pre-training and supervised training, we use an input sequence and a target sequence. 
For the bimodal model, we concatenate a sequence of natural language text and the corresponding sequence of programming language text as an input. For the unimodal model, we simply use each coded function as an input sequence. During self-supervised training, spans of the input sequence are randomly masked and the target sequence \cite{DBLP:journals/corr/abs-1910-10683} is formed as the concatenation of the same sentinel tokens and the real masked spans/tokens.

\subsection{Model Architecture}
CoTexT follows the sequence-to-sequence encoder-decoder architecture proposed by \cite{DBLP:journals/corr/VaswaniSPUJGKP17}. We initialize the Base T5 model released by \cite{DBLP:journals/corr/abs-1910-10683} which has 220 million parameters. We train the model with a 0.001 learning rate and an input/target length of 1024. With the provided TPU v2-8 on Google Colab, we train with the recommended setting of model parallelism 2 and batch size 128.

\subsection{Multi-task Learning}
The model is trained with maximum likelihood objective (that is using "teacher forcing" \cite{10.1162/neco.1989.1.2.270}) regardless of the text-code or code-text tasks. Therefore, for CoTexT, we leverage the potential for Multi-Task learning \cite{DBLP:journals/corr/abs-1910-10683} to complete both text-code and code-text generation on CodeSummarization and Code Refinement tasks. To specify the task our model should perform, we simply add a task-specific prefix to the input sequence. For example, when fine-tuning of the CodeSummarization task for each programming language, we simply prepend a prefix for each PL name (i.e., Java) to the input sequence.

\begin{figure}
    \centering
    \includegraphics[width=0.5\textwidth,height=\textheight,keepaspectratio]{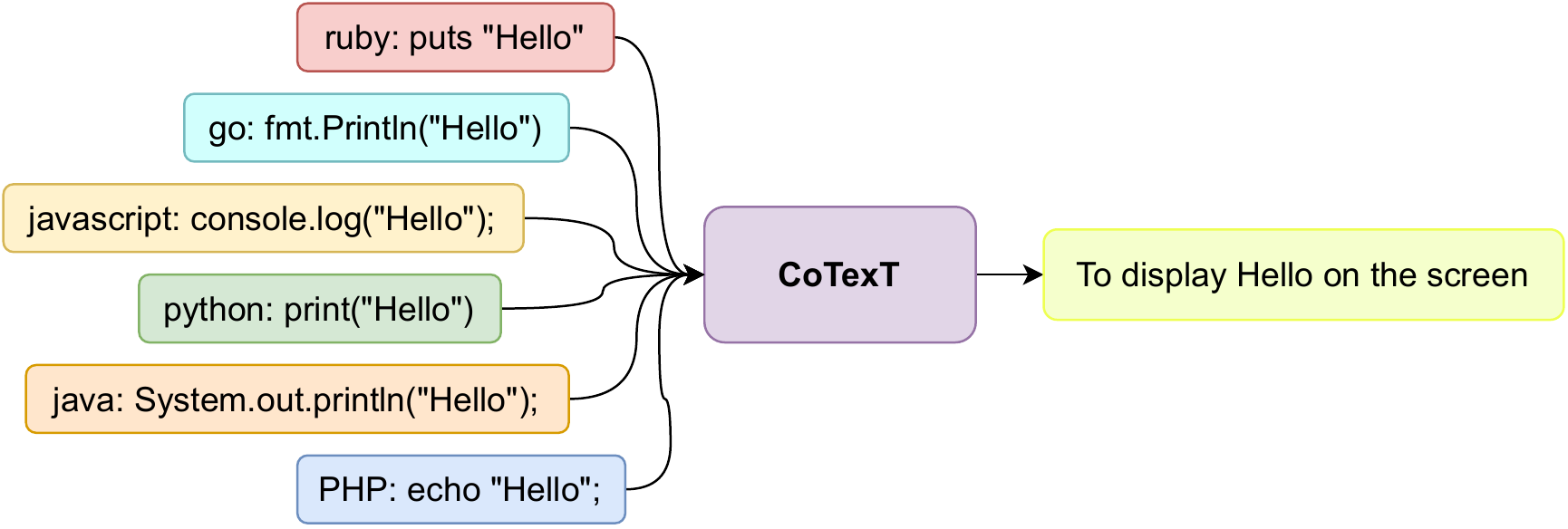}
    \caption{An illustration about Multi-task learning}
    \label{fig:multitask}
\end{figure}

\section{Experiments}

In this section, we will first describe the benchmark dataset for code intelligence CodeXGLUE, then we will explain the experimental setup on the tasks we perform and discuss the results of each task. The evaluation datasets are summarized in Table \ref{statistic_table}.

\subsection{CodeXGLUE}
General Language Understanding Evaluation benchmark for CODE (CodeXGLUE) \cite{lu2021codexglue} is a benchmark dataset to facilitate machine learning studies on code understanding and code generation problems. This dataset includes a collection of code intelligence tasks (both classification and generation), a platform for model evaluation, and a leaderboard for comparison. CodeXGLUE has 10 code intelligence tasks including code-text, text-code, code-code, and text-text scenarios. For CoTexT, we focus on Code Summarization, Code Generation, Code Refinement, and Defect Detection tasks. 

\begin{table*}[]
\caption{The input and target sequence length settings for each self-supervised learning, code summarization, code generation, code refinement, and defect detection task}
\begin{tabular}{@{}lllll@{}}
\toprule
Task                                      & Dataset                        & Task Type                   & Input Length         & Target Length        \\ \midrule
\multirow{2}{*}{Self-supervised Learning} & CodSearchNet Corpus            &                             & 1024                 & 1024                 \\
                                          & GitHub Repositories            &                             & 1024                 & 1024                 \\ \midrule
Code Summarization     & CodeSearchNet & Multi-Task & 512 & 512 \\ \midrule
Code Generation                           & CONCODE                        & Single-Task                 & 256                  & 256                  \\ \midrule
Code Refinement                           & Bugs2Fix\textsubscript{small}               & \multirow{2}{*}{Multi-Task} & \multirow{2}{*}{512} & \multirow{2}{*}{512} \\
                                          & Bugs2Fix\textsubscript{medium}               &                             &                      &                      \\ \midrule
Defect Detection                          & Devign                         & Single-Task                 & 1024                 & 5                    \\ \bottomrule
\end{tabular}
\label{setting}
\end{table*}
\subsection{Evaluation Tasks}
We evaluate our programming language and natural language generation tasks on TPU v2-8 with the settings from the original T5 model \cite{DBLP:journals/corr/abs-1910-10683}. The input length and target length for each task are described in Table \ref{setting}.
\subsubsection{Code Summarization}
For Code Summarization, the objective is to generate a natural language description for a given code snippet. The task includes a CodeSearchNet dataset \cite{DBLP:journals/corr/abs-1909-09436} with 6 different programming languages: Python, Java, Javascript, PHP, Ruby, Go. The data comes from public open-source non-fork GitHub repositories and the annotations are extracted from function documentation as described in \cite{DBLP:journals/corr/abs-1909-09436}.

\subsubsection{Code Generation}
Text-to-Code Generation aims to generate a coded function given a natural language description. This task is completed using the CONCODE dataset \cite{DBLP:journals/corr/abs-1808-09588}, a well-known dataset for Java language generation. Within the dataset, there are tuples which contain a natural language description, code environments, and code snippets. The goal is to generate the correct Java function from the natural language description in the form of Javadoc-style method comments.

\subsubsection{Code Refinement}
Code Refinement, or Code Repair, aims to automatically correct bugs in Java code. We used the Bug2Fix corpus released by CodeXGLUE \cite{lu2021codexglue}, which divides the task into 2 subsets: SMALL and MEDIUM The small dataset includes only Java code functions with fewer than 50 tokens. The medium dataset includes functions with 50-100 tokens.

\subsubsection{Defect Detection}
For Defect Detection tasks, we attempt to classify whether a PL snippet contains vulnerabilities that could lead to damaging outcomes such as resource leaks or DoS attacks. The task uses the Devign dataset \cite{zhou2019devign}, which contains C programming language from open-source projects. This dataset is labeled based on security-related commits. For details on the annotation process, refer to \cite{zhou2019devign}.

\def\arraystretch{1.2}%
\setlength{\tabcolsep}{0.9em} %
\begin{table*}[!htbp]
\centering
\caption{Data statistics about Code Intelligence datasets}
\begin{tabular}{p{1.8cm}|P{3.2cm}|P{2.5cm}|P{0.7cm}|P{0.7cm}|P{0.7cm}|p{1.3cm}}
\hline
\multirow{2}{1.8cm}{\textbf{Category}} & \multirow{2}{3.2cm}{\centering\textbf{Task}} & \multirow{2}{2.5cm}{\centering\textbf{Dataset}} & \multicolumn{3}{c|}{\textbf{Size}}  & \multirow{2}{1.77cm}{\textbf{Language}} \\

\cline{4-6}
 & & & Train & Val & Test &  \\
\hline
\multirow{6}{1.8cm}{Code-Text} & \multirow{6}{3.2cm}{\centering Code Summarization \cite{lu2021codexglue}} & \multirow{6}{2.5cm}{CodeSearchNet} & 164K & 5.1K & 10.9K & Java \\
 &  &  & 58K & 3.8K & 3.2K & Javascript \\
 &  &   & 251K & 13.9K & 14.9K & Python \\
&  &  & 241K & 12.9K & 14K & PHP \\
 & &  & 167K & 7.3K & 8.1K & Go \\
 &  &  & 24K & 1.4K & 1.2K & Ruby \\
\hline
\multirow{4}{1.8cm}{Code-Code}  & Defect Detection & \multirow{2}{2.5cm}{\centering Devign} & \multirow{2}{0.7cm}{21K} & \multirow{2}{0.7cm}{2.7K} & \multirow{2}{0.7cm}{2.7K} & \multirow{2}{1.77cm}{C} \\
  & \cite{zhou2019devign} &  &  &  & & \\
\cline{2-7}
& \multirow{2}{3.2cm}{\centering Code Refinement \cite{lu2021codexglue}}& Bugs2Fix\textsubscript{small} & 46K & 5.8K & 5.8K & \multirow{2}{1.77cm}{Java} \\
\cline{3-6}
&  & Bugs2Fix\textsubscript{medium} & 52K & 6.5K & 6.5K &  \\
\hline
\multirow{2}{1.8cm}{Text-Code} & Code Generation & \multirow{2}{2.5cm}{\centering CONCODE} & \multirow{2}{0.7cm}{100K} & \multirow{2}{0.7cm}{2K} & \multirow{2}{0.7cm}{2K} & \multirow{2}{1.77cm}{Java} \\
 & \cite{DBLP:journals/corr/abs-1808-09588} &  &  &  &  &  \\
\hline
\end{tabular}
\label{statistic_table}

\end{table*}

\subsection{Experimental Setup}
\subsubsection{Baselines}
We compare our model with some well-known pre-trained models:
\begin{itemize}
    \item CodeGPT, CodeGPT-adapted are based on the architecture and training objective of GPT-2 \cite{DBLP:journals/corr/abs-1907-05774}. CodeGPT is pre-trained from scratch on CodeSearchNet dataset \cite{lu2021codexglue} while CodeGPT-adapted learns this dataset starting from the GPT-2 checkpoint. 
    \item CodeBERT \cite{feng-etal-2020-codebert} employs the same architecture as RoBERTa \cite{liu2020roberta} but aims to minimize the combined loss from masked language modeling and replaced token detection.
    \item PLBART \cite{ahmad2020summarization} is a Transformer-based model. BART \cite{lewis-etal-2020-bart} is trained on PL corpora using three learning strategies: token masking, token deletion, and token infilling.
\end{itemize}

\subsubsection{Performance Metrics}
\begin{itemize}
    \item BLEU \cite{10.3115/1073083.1073135} is an algorithm which performs automatic evaluation of machine-translated text. This method calculates  the n-gram similarity of a candidate translation compared to a set of reference texts. Similar to \cite{feng-etal-2020-codebert} and \cite{ahmad2020summarization}, we use smooth BLEU-4 score \cite{lin-och-2004-orange} for Code Summarization and corpus-level BLEU score for all remaining tasks.
    \item CodeBLEU \cite{ren2020codebleu} is designed to consider syntactic and semantic features of codes based on the abstract syntax tree and the data flow structure.
    \item Accuracy is the ratio of the number of generated sequences that harmonise the reference to the total number of observations.
\end{itemize}

\begin{table*}[!htbp]
\centering
\caption{Test result on Code Generation task}

\begin{tabular}{l|l|l|ll}
\cline{1-4}
\multicolumn{1}{l|}{\multirow{2}{*}{\textbf{Model}}} & \multicolumn{3}{c}{\textbf{Text2Code Generation}} &  \\
\cline{2-4}
\multicolumn{1}{c|}{}                       & EM          & BLEU        & CodeBLEU      &  \\ \cline{1-4}
PLBART                                      & 18.75       & {\ul 36.69}       & 38.52         &  \\
CodeGPT-adapted                             & 20.10        & 32.79       & 35.98         &  \\
CodeGPT                                     & 18.25       & 28.69       & 32.71         &  \\
T5                                          & 18.65       & 32.74       & 35.95         &  \\ \cline{1-4}
CoText (1-CCG)       & 19.45       & 35.40       & 38.47         &  \\
CoText (2-CC)             & {\ul 20.10}  & 36.51       & {\ul 39.49}         &  \\
CoText (1-CC)            & \textbf{20.10} & \textbf{37.40}       & \textbf{40.14}         &  \\ \cline{1-4}
\end{tabular}
\begin{tablenotes}
      \small
      \item \textit{Notes:} The best scores are in bold and second best scores are underlined. The baseline scores were obtained from the CodeXGLUE's Leaderboard (https://microsoft.github.io/CodeXGLUE/)
\end{tablenotes}
\label{code_generation}
\end{table*}
\begin{table*}[!htbp]
\centering
\caption{Test result on Code Summarization task}
\begin{tabular}{p{3.5cm}|P{0.8cm}|P{0.8cm}|P{1.6cm}|P{0.8cm}|P{1.1cm}|p{0.8cm}|p{0.8cm}}
\toprule
\textbf{Model}                                     & \textbf{All}            & \textbf{Ruby}           & \textbf{Javascript}     & \textbf{Go}             & \textbf{Python}         & \textbf{Java}          & \textbf{PHP}            \\ \midrule
RoBERTa                                   & 16.57          & 11.17          & 11.90          & 17.72          & 18.14          & 16.47         & 24.02          \\
CodeBERT                                  & 17.83          & 12.16          & 14.90          & 18.07          & 19.06          & 17.65         & \textbf{25.16} \\
PLBART                                    & 18.32          & \textbf{14.11} & \textbf{15.56} & 18.91          & 19.3           & 18.45         & 23.58          \\
T5                                    & 18.35          & 14.18 & 14.57 & {\ul 19.17}          & 19.26           & 18.35         & 24.59          \\
\midrule
CoTexT (1-CCG) & 18.00          & 13.23          & 14.75          & 18.95    & 19.35          & 18.75         & 22.97          \\
CoTexT (2-CC)        & {\ul 18.38}    & 13.07          & 14.77          & \textbf{19.37} & {\ul 19.52}    & \textbf{19.1} & 24.47          \\
CoTexT (1-CC)       & \textbf{18.55} & {\ul 14.02}    & {\ul 14.96}    & 18.86          & \textbf{19.73} & {\ul 19.06}   & {\ul 24.58}    \\ \bottomrule
\end{tabular}
\begin{tablenotes}
      \small
      \item \textit{Notes:} The best scores are in bold and second best scores are underlined. The baseline scores were obtained from the CodeXGLUE's Leaderboard (https://microsoft.github.io/CodeXGLUE/)
\end{tablenotes}
\label{code_summarization}
\end{table*}

\begin{table*}[!htbp]
\centering
\caption{Test result on Code Refinement task}
\begin{tabular}{p{3.5cm}|P{0.9cm}|P{1cm}|P{1.6cm}|P{0.9cm}|P{1cm}|p{1.6cm}}
\toprule
                                      & \multicolumn{3}{c|}{\textbf{Small test set}}               & \multicolumn{3}{c}{\textbf{Medium test set}}              \\
\cline{2-7}
\textbf{Model}                                 & BLEU           & Acc(\%)        & CodeBLEU       & BLEU           & Acc(\%)        & CodeBLEU       \\ \midrule
Transformer                           & 77.21          & 14.70          & 73.31          & {\ul 89.25}    & 3.70           & 81.72          \\
CodeBERT                              & 77.42          & 16.40          & 75.58          & \textbf{91.07} & 5.16           & \textbf{87.52} \\
PLBART                                & 77.02          & 19.21          & /              & 88.5           & 8.98           & /              \\
T5                                    & 74.94          & 15.3           & 75.85          & 88.28          & 4.11           & 85.61          \\ \midrule
CoTexT (1-CCG) & 76.87          & 20.39          & {\ul 77.34}    & 88.58          & 12.88          & {\ul 86.05}    \\
CoTexT (2-CC)        & {\ul 77.28}    & \textbf{21.58} & \textbf{77.38} & 88.68          & {\ul 13.03}    & 84.41          \\
CoTexT (1-CC)       & \textbf{77.79} & {\ul 21.03}    & 76.15          & 88.4           & \textbf{13.11} & 85.83          \\ \bottomrule
\end{tabular}
\begin{tablenotes}
      \small
      \item \textit{Notes:} The best scores are in bold and second best scores are underlined. The baseline scores were obtained from the CodeXGLUE's Leaderboard (https://microsoft.github.io/CodeXGLUE/)
\end{tablenotes}
\label{code_refinement}
\end{table*}

\begin{table}[!htbp]
\centering
\caption{Test result on Defect Detection task}
\begin{tabular}{p{4cm}|P{1.5cm}}
\toprule
\textbf{Model}                                & \textbf{Accuracy}       \\ \midrule
RoBERTa                              & 61.05          \\
CodeBERT                             & 62.08          \\
PLBART                               & 63.18          \\
T5                                   & 61.93          \\ \midrule
CoTexT (1-CCG) & \textbf{66.62} \\
CoTexT (2-CC)       & 64.49          \\
CoTexT (1-CC)      & {\ul 65.99}    \\ \bottomrule
\end{tabular}
\label{defect_detection}
\begin{tablenotes}
      \small
      \item \textit{Notes:} The best scores are in bold and second best scores are underlined. The baseline scores were obtained from the CodeXGLUE's Leaderboard (https://microsoft.github.io/CodeXGLUE/)
\end{tablenotes}
\end{table}

\section{Results}

\subsection{Multi-Task Learning}
We first report the result of CoTexT in Multi-Task Learning tasks including Code Summarization and Code Refinement.

\subsubsection{Code Summarization}
For the Code Summarization task, we perform Multi-Task Learning by using the T5 framework \cite{DBLP:journals/corr/abs-1910-10683} to finetune CoTexT on 6 diferent programming language (Ruby, Javascript, Go, Python, Java, and PHP). The results of the Code Summarization task are shown in Table \ref{code_summarization}.

First, we observe that the base T5, which is pre-trained only on the general domain corpus (C4), is effective on this task. In fact, base T5 achieves higher overall results on the BLEU-4 metric compared to all other related models on the CodeXGLUE leaderboard. This shows the importance of domain-specific T5 models, which we expect to achieve superior results compared to base T5. 

We further observe that CoTexT achieves state-of-the-art (SOTA) on the overall score, the Python-specific score, the Java-specific score, and the Go-specific score. While CoTexT does not significantly outperform other pre-trained models, we observe that CoTexT achieves SOTA on two very common programming languages (Python and Java) while still obtaining competitive results on other programming languages. We attribute this result to the large amount of training data for Python and Java compared to the other languages (training size described in Table \ref{statistic_table}). Based on this result, CoTeXT has the potential to further surpass competitor models as more training data becomes availible.

\subsubsection{Code Refinement}
We also tested CoTexT by performing multi-task learning for Code Refinement. In this case, both the small and medium test sets have a task registry with respective prefix prepending to the input sequence.

The Code Refinement results of each model are shown in Table \ref{code_refinement}. For this task, the base T5, which is pre-trained only on natural language text, does not perform well compared to other transformer-based models. Yet, after the training on a large programming language corpus, the result from CoTexT improves significantly on all metrics for both small and medium test sets. CoTexT achieves SOTA for all metrics on the small test set and on the accuracy metric for the medium test set. 

\subsection{Single-Task Learning}
In addition to multi-task learning, we also evaluate CoTexT performance single-task learning with a Code Generation Task and a classification task relating to Defect Detection.

\subsubsection{Code Generation}
In Table \ref{code_generation}, we reported our results for the Code Generation task wherein natural language is translated into Java code. The result shows that our proposed model achieves SOTA results based on 3 metrics: Exact Match (EM), BLEU, and CodeBLEU. For each individual metric, CoTexT has only slightly outperformed other models (e.g both CoTexT and CodeGPT-adapted achieve 20.10 for EM). However, our model is consistently superior across the 3 metrics. Prior to CoTexT, CodeGPT-adapted was SOTA for the EM metric and PLBART was SOTA for the BLUE/CodeBLUE metrics. From this result, we infer that CoTexT has the best overall performance on this task and has great potential in the area of code generation.  

\subsubsection{Defect Detection}

The Defect Detection results are shown in Table \ref{defect_detection}. Specifically, CoText outperforms the previous SOTA model (PLBART) by 3.44\%. For this task, extra training on a large programming corpus allows CoTexT to outperform all other models and achieve SOTA results. The Defect Detection dataset consists of code written in the C programming language, which is not contained in our training data. Our model has a strong understanding of similar languages, and is thus able to  perform Defect Detection in C with improved results compared to competitor models.

\section{Conclusion}
In this manuscript, we introduced CoTexT, a pre-trained language representation for both programming language and natural language. CoTexT focused on text-code and code-text understanding and generating. Leveraging  the T5 framework \cite{DBLP:journals/corr/abs-1910-10683}, we showed that pre-training on a large programming language corpus is effective for a diverse array of tasks within the natural language and programming language domain. CoTexT achieves state-of-the-art results on 4 CodeXGLUE code intelligence tasks: Code Summarization, Code Generation, Code Refinement, and Code Detection. For future work, we plan to test CoTexT on a broader range of programming language and natural language generation tasks, such as autocompletion or code translation.

\bibliographystyle{acl_natbib}
\bibliography{acl2021}


\end{document}